\documentclass{article}
\usepackage{ijcai24}

\usepackage{times}
\usepackage{soul}
\usepackage{url}
\usepackage[hidelinks]{hyperref}
\usepackage[utf8]{inputenc}
\usepackage[small]{caption}
\usepackage{graphicx}
\usepackage{amsmath}
\usepackage{amsthm}
\newtheorem{definition}{Definition}
\usepackage{amssymb}
\usepackage{booktabs}
\usepackage{comment}
\usepackage{subfig}
\usepackage{algorithm}
\usepackage{algorithmic}
\usepackage[switch]{lineno}
\usepackage{multirow}

\title{Hypergraph Self-supervised Learning with Sampling-efficient Signals}

\iftrue
\author{
Fan Li$^1$\and
Xiaoyang Wang$^{1}$\footnote{Corresponding author}\and
Dawei Cheng$^{2}$\and
Wenjie Zhang$^1$\and
Ying Zhang$^3$\And
Xuemin Lin$^4$\\
\affiliations
$^1$University of New South Wales, Sydney, Australia\\
$^2$Tongji University, Shanghai, China\\
$^3$University of Technology Sydney, Sydney, Australia\\
$^4$Shanghai Jiaotong University, Shanghai, China\\
\emails
\{fan.li8, xiaoyang.wang1,wenjie.zhang\}@unsw.edu.au,
dcheng@tongji.edu.cn,
ying.zhang@uts.edu.au,
xuemin.lin@gmail.com
}
\fi

\begin{document}

\maketitle

\begin{abstract}
Self-supervised learning (SSL) provides a promising alternative for representation learning on hypergraphs without costly labels. However, existing hypergraph SSL models are mostly based on contrastive methods with the instance-level discrimination strategy, suffering from two significant limitations: (1) They select negative samples arbitrarily, which is unreliable in deciding similar and dissimilar pairs, causing training bias. (2) They often require a large number of negative samples, resulting in expensive computational costs. To address the above issues, we propose SE-HSSL, a hypergraph SSL framework with three sampling-efficient self-supervised signals. Specifically, we introduce two sampling-free objectives leveraging the canonical correlation analysis as the node-level and group-level self-supervised signals. Additionally, we develop a novel hierarchical membership-level contrast objective motivated by the cascading overlap relationship in hypergraphs, which can further reduce membership sampling bias and improve the efficiency of sample utilization. Through comprehensive experiments on 7 real-world hypergraphs, we demonstrate the superiority of our approach over the state-of-the-art method in terms of both effectiveness and efficiency.
\end{abstract}

\section{Introduction}

The hypergraph, where each hyperedge can connect any number of nodes, is a generalization data structure of the simple pairwise graph. It provides a natural and expressive way to model complex high-order relationships among entities in diverse real-world applications such as recommender systems~\cite{xia2022self}, computer vision~\cite{yu2012adaptive}, and neuroscience~\cite{xiao2019multi}. 

In recent years, hypergraph representation learning has attracted increasing attention from both academia and industry to deal with hypergraph data~\cite{yi2020hypergraph,jia2021hypergraph,antelmi2023survey}. Hypergraph neural networks (HGNNs) have become a popular tool in this domain, demonstrating remarkable effectiveness~\cite{Yifan:19,chien2021you}. However, task-specific labels can be extremely scarce in hypergraph datasets~\cite{wei2022augmentations}, and deep models are prone to overfitting when trained with sparse supervised signals~\cite{thakoor2021bootstrapped}. Therefore, it is crucial yet challenging to explore self-supervised learning for HGNNs, where only limited or even
no labels are needed.

Current hypergraph SSL models are mostly contrastive-based, which aim to maximize the agreement between two augmented views derived from the original hypergraph. Previous studies, such as HCCF~\cite{xia2022hypergraph} and $S^{2}$-HHGR~\cite{zhang2021double}, employ contrastive strategies to mitigate label scarcity in hypergraph-based recommender systems, but they are not general methods designed for hypergraph contrastive learning (HCL). HyperGCL~\cite{wei2022augmentations}, as the first comprehensive hypergraph SSL method, focuses on generating augmented views that better preserve high-order relations compared to fabricated augmentations. TriCL~\cite{lee2023m} proposes tri-directional contrast objectives in hypergraphs to capture multi-view structural information, achieving state-of-the-art performance. 

Nevertheless, existing contrastive-based approaches suffer from two limitations: (1) They employ the instance-level discrimination strategy, that is, pulling together the representation of the same node/hyperedge in the two augmented hypergraphs while pushing apart every other node/hyperedge pairs. However, this approach is unreliable for determining similar/dissimilar pairs. For example, consider a co-author hypergraph, papers (nodes) published by the same author (hyperedge) are more likely to have similar topics. Authors who co-publish (overlap) a significant number of papers may share similar research interests. Treating them as negative pairs arbitrarily would introduce training bias, resulting in less discriminative representations. (2) They require a large number of negative pairs to achieve the best performance. However, generating a large number of negative pairs often comes with prohibitive computational costs, particularly for large hypergraphs. In TriCL, in addition to node-level and group-level contrast, it introduces membership-level contrast to maximize the agreements between each group (hyperedge) and its members (nodes) in two augmented views. Although this signal has been proven to be highly effective, it requires a large number of node-hyperedge negative pairs to achieve the SOTA performance, leading to a complexity of $O(|\mathcal{V}| \times |\mathcal{E}|)$ in scoring function computation, where $\mathcal{V}$ and $\mathcal{E}$ denote the number of nodes and hyperedges, respectively. This would greatly limit the training speed. Moreover, it also follows the instance-level contrast strategy when handling positive/negative membership pairs, which can introduce the training bias mentioned above. 

To address the above issues, we introduce SE-HSSL, a sampling-efficient framework for hypergraph self-supervised learning. Specifically, it first employs an HGNN to encode the original hypergraph and two augmented views, generating multi-view node and hyperedge representations. Then, to capture high-order relations in hypergraphs and reduce sampling bias caused by previous contrastive-based methods, we propose tri-directional sampling-efficient self-supervised signals. For node-level and group-level learning, we design two optimization objectives based on the idea of canonical correlation analysis (CCA)~\cite{andrew2013deep}. These two signals allow the model to maximize the agreement between two augmented views without reliance on any negative sample. Furthermore, instead of relying on instance-level membership discrimination, which focuses on distinguishing between ``real'' and ``fake'' node-hyperedge memberships, we present a novel hierarchical membership-level contrast signal based on the overlap structure within hypergraphs. This design effectively addresses sampling bias in membership-level learning and substantially reduces the number of negative pairs, leading to improved effectiveness and efficiency. Finally, the model jointly optimizes the three objectives, facilitating comprehensive representation learning in hypergraphs.
Our main contributions are summarized as follows:

\begin{itemize}
    \item We propose SE-HSSL, a hypergraph self-supervised learning method that can address training bias and sampling inefficiency challenges in existing HCL methods.
    \item In our HSSL framework, we introduce sampling-free CCA-based node- and group-level objectives to reduce training bias and generate more discriminative representations. Furthermore, our novel hierarchical membership contrast signal can effectively preserve membership relations with only a few sampled node-hyperedge pairs.
    \item We conduct a comprehensive evaluation of SE-HSSL on 7 real-world datasets. The experimental results demonstrate its superiority over strong baselines across various benchmark downstream tasks. The efficiency test shows that it can achieve at least 2x speedup during
    training compared to the state-of-the-art on most datasets.
\end{itemize}


\section{Preliminary}

\paragraph{Notations.} Let $\mathcal{H}(\mathcal{V},\mathcal{E})$ represent a   hypergraph with vertex set $\mathcal{V}$ = $\{v_{i}\}_{i=1}^{|\mathcal{V}|}$ and hyperedge set $\mathcal{E}$ = $\{e_{j}\}_{j=1}^{|\mathcal{E}|}$. $\mathbf{X} \in \mathbb{R}^{|\mathcal{V}| \times F}$ denotes the $F$ dimensional node feature matrix. As each hyperedge $e \in \mathcal{E}$ is a subset of $\mathcal{V}$, we denote $|e|$ as the size of $e$. Each hyperedge $e_{j}$ is associated with a positive number $w_{j}$ as hyperedge weight, and all the weights formulate a diagonal matrix $\mathbf{W} \in \mathbb{R}^{|\mathcal{E}| \times |\mathcal{E}|}$. The hypergraph structure can be represented by an incidence matrix $\mathbf{H} \in \mathbb{R}^{|\mathcal{V}| \times |\mathcal{E}|}$, where each entry $h_{ij}=1$ if $v_{i} \in e_{j}$ and $h_{ij}=0$ otherwise. We represent the degree of vertices using the diagonal matrix $\mathbf{D}_{v} \in \mathbb{R}^{|\mathcal{V}| \times |\mathcal{V}|}$, where each entry is $d(v_{i})=\sum_{e_{j} \in \mathcal{E}}w_{j} \cdot h_{ij}$. The degree of hyperedges is denoted by the diagonal matrix
$\mathbf{D}_{e} \in \mathbb{R}^{|\mathcal{E}| \times |\mathcal{E}|}$, with each element $\delta(e_{j})=\sum_{v_{i} \in e_{j}}h_{ij}$ representing the number of nodes connected by $e_{j}$.


\paragraph{Hypergraph Neural Networks (HGNNs).} HGNNs apply a two-stage neighborhood aggregation scheme to learn hypergraph representations. This entails updating the hyperedge representation by aggregating representations of its incident nodes, and updating the node representation by propagating information from representations of its incident hyperedges:
\begin{equation}
    \begin{split}
        z_{e,j}^{(l)} & = f_{\mathcal{V} \rightarrow \mathcal{E}}^{(l)}(z_{e,j}^{(l-1)},\{z_{v,k}^{(l-1)}|v_{k} \in e_{j} \}) \\
        z_{v,i}^{(l)} & = f_{\mathcal{E} \rightarrow \mathcal{V}}^{(l)}(z_{v,i}^{(l-1)},\{z_{e,k}^{(l)}|v_{i} \in e_{k} \}) \\
    \end{split}
\end{equation}
where $z_{e,j}^{(l)}, z_{v,i}^{(l)}$ are embeddings of $e_{j}$ and $v_{i}$ at layer $l$, respectively. $f_{\mathcal{V} \rightarrow \mathcal{E}}^{(l)}$ and $f_{\mathcal{E} \rightarrow \mathcal{V}}^{(l)}$ are two permutation-invariant functions
which aggregates information from nodes and hyperedges,
respectively, at the $l$-th layer. 
Recent works have proposed different $f_{\mathcal{V} \rightarrow \mathcal{E}}^{(l)}$ and $f_{\mathcal{E} \rightarrow \mathcal{V}}^{(l)}$, leading to multiple HGNN variants~\cite{kim2020hypergraph,chien2021you}.

\paragraph{Deep Canonical Correlation Analysis (DCCA).} CCA is a classical multivariate analysis method designed to identify the linear combinations of variables from each set that are maximally correlated with each other. Recent research has explored the application of CCA in multi-view learning by leveraging deep neural networks instead of linear transformations~\cite{andrew2013deep,jing2014intra}. This technique helps capture complex and nonlinear relationships that are commonly shared between two sets of multidimensional vectors. Assuming $X_{1}$ and $X_{2}$ are input data from two views, the DCCA optimization objective can be formulated as:
\begin{equation}
    \begin{split}
         \underset{\theta_{1},\theta_{2}}{\min} & \mathcal{L}_{C}(f_{\theta_{1}}(X_{1}),f_{\theta_{2}}(X_{2})) \\
         & +\lambda(L_{D}(f_{\theta_{1}}(X_{1}))+L_{D}(f_{\theta_{2}}(X_{2})))
    \end{split}
\end{equation}
where $f_{\theta_{1}},f_{\theta_{2}}$ are two neural networks, and $\lambda$ is a Lagrangian multiplier. $L_{C}$ is the invariance term that measures the correlation between the two views, while $L_{D}$ is the feature-level decorrelation term. Following the popular setting in soft CCA~\cite{chang2018scalable}, $L_{D}$ computes the $L_{1}$ distance between $f_{\theta_{i}}(X_{i})$ and identity matrix, for $i=1,2$. 


\paragraph{Problem statement.} Given a hypergraph $\mathcal{H}(\mathbf{H},\mathbf{X})$, 
our objective is to train an optimal hypergraph encoder $f_{\theta}: (\mathbf{H},\mathbf{X}) \rightarrow (\mathbf{Z}_{v},\mathbf{Z}_{e})$ in an unsupervised manner. This encoder maps the node and hyperedge to low-dimensional embeddings, which can be beneficial for various downstream tasks, such as node classification and node clustering.

\section{Methodology}

\begin{figure*}[t]
\centering
\includegraphics[width=0.85\textwidth]{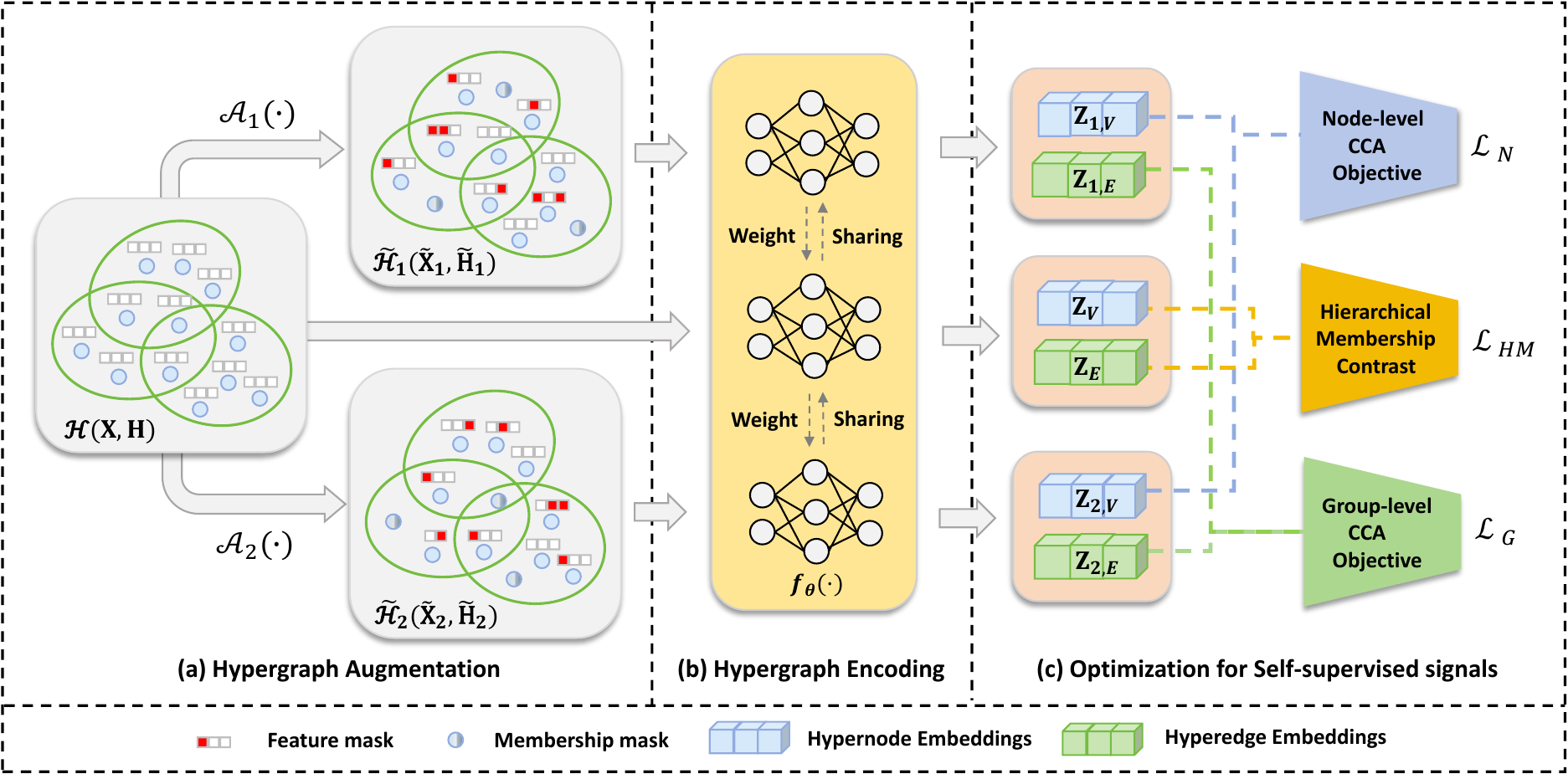} 
\caption{The architecture of the hypergraph self-supervised learning framework SE-HSSL.}
\vspace{-9pt}
\label{fig:arch}
\end{figure*}

In this section, we introduce our proposed Hypergraph SSL framework. We begin by briefly presenting an overview of SE-HSSL. Then, we bring forward the details of its three
modules. Finally, we describe the joint optimization strategy. 

\subsection{Overview}

As illustrated in Figure~\ref{fig:arch}, our hypergraph SSL architecture consists of the following three major components: hypergraph augmentation, hypergraph encoding, and optimization for self-supervised signals. Initially, the given input hypergraph $\mathcal{H}(\mathbf{H},\mathbf{X})$ undergoes data augmentation to obtain two correlated views $\Tilde{\mathcal{H}}_{1}(\Tilde{\mathbf{H}}_{1},\Tilde{\mathbf{X}}_{1})$ and $\Tilde{\mathcal{H}}_{2}(\Tilde{\mathbf{H}}_{2},\Tilde{\mathbf{X}}_{2})$. Next, we utilize a shared HGNN encoder $f_{\theta}$ to generate node and hyperedge representations for the original hypergraph $\mathcal{H}$ and the two augmented views, $\Tilde{\mathcal{H}}_{1}$ and $\Tilde{\mathcal{H}}_{2}$. 
While most existing HSSL approaches apply instance-level contrast signals that could lead to sampling bias and heavy computation burden, in this paper,
we construct three novel sampling-efficient self-supervised signals: a node-level CCA objective, a group-level CCA objective, and a hierarchical membership-level contrast objective. Lastly, we jointly optimize the loss function of these signals, denoted as $\mathcal{L}_{N}$,$\mathcal{L}_{G}$, and $\mathcal{L}_{HM}$, respectively.

\subsection{Hypergraph Augmentation}

We consider two manually designed hypergraph augmentation strategies: node feature masking and membership masking. In node feature masking, we first sample a random vector $\Tilde{m} \in \{0,1\}^{F}$, where each dimension is independently drawn from a Bernoulli distribution, i.e., $\Tilde{m}_{j} \sim \mathcal{B}(1-p^{f}), \forall i$. The augmented feature matrix $\Tilde{\mathrm{X}}$ is then computed as follows:
\begin{equation}
    \Tilde{\mathrm{X}}=[x_{1} \odot \Tilde{m};x_{2} \odot \Tilde{m},...,x_{|\mathcal{V}|} \odot \Tilde{m}]^{T}
\end{equation}
where $\odot$ denotes the element-wise multiplication and $[\cdot ; \cdot]$ is the concatenate operator. Regarding membership masking, we aim to disrupt high-order relations in hypergraphs by randomly kicking out vertices from hyperedges. We construct a masking matrix $\mathbf{M} \in \{0,1\}^{|\mathcal{V}| \times |\mathcal{E}|}$ where each entry $m_{ij} \sim \mathcal{B}(1-p^{m})$. The membership masking augmentation on hypergraph topology can be formulated as: 
\begin{equation}
    \Tilde{\mathbf{H}}= \mathbf{M} \odot \mathbf{H} 
\end{equation}
$p^{f}$ and $p^{m}$ denote drop probability for node features and node-hyperedge membership links, respectively.

\subsection{Hypergraph Encoder}

Following the classic hypergraph message passing scheme, we take the HGNN with the element-wise mean pooling layer as our hypergraph encoder. Formally, the $l$-th layer of HGNN can be represented as:
\begin{equation}
    \begin{split}
        \mathbf{Z}_{\mathcal{E}}^{(l)} & = \phi(\mathbf{D}_{e}^{-1}\mathbf{H}^{T}\mathbf{Z}_{\mathcal{V}}^{(l-1)}\mathbf{\Theta}_{\mathcal{E}}^{(l)})\\
        \mathbf{Z}_{\mathcal{V}}^{(l)} & =\phi(\mathbf{D}_{v}^{-1}\mathbf{H}\mathbf{W}\mathbf{Z}_{\mathcal{E}}^{(l)}\mathbf{\Theta}_{\mathcal{V}}^{(l)})\\ 
    \end{split}
\end{equation}
where $\mathbf{Z}_{\mathcal{E}}^{(l)} \in \mathbb{R}^{|\mathcal{V}| \times D},\mathbf{Z}_{\mathcal{V}}^{(l)} \in \mathbb{R}^{|\mathcal{E}| \times D}$ are hyperedge and node representations at the $l$-th layer. $D$ is the embedding dimensionality. $\mathbf{\Theta}_{\mathcal{E}}^{(l)}$ and $\mathbf{\Theta}_{\mathcal{V}}^{(l)}$ denote trainable parameters for $f_{\mathcal{V} \rightarrow \mathcal{E}}^{(l)}$ and $f_{\mathcal{E} \rightarrow \mathcal{V}}^{(l)}$, respectively. $\phi(\cdot)$ represents PReLU nonlinear activation function. $\mathbf{W}$ is initialized as an identity matrix, which means equal weights for all hyperedges.

\subsection{Self-supervised Signal Construction}

To address the issues of training bias and heavy computation burden caused by arbitrary negative sampling, we propose tri-directional sampling-efficient self-supervised signals. 

\paragraph{Node-level CCA objective.} Most contrastive methods rely on negative pairs to avoid degenerated solutions in feature space~\cite{zhang2021canonical}, that is, different dimensions are highly correlated and can capture the same information, making learned representations less discriminative. Following the DCCA, we can maximize the agreement of two augmented views with the invariance term and prevent degenerated solutions with the feature-level decorrelation term. In this way, we are not required to select negative pairs arbitrarily, thereby reducing sampling bias.
Specifically, we first normalize node embedding matrix $\mathbf{Z}_{\mathcal{V}}$ along instance dimension so that each feature dimension follows a distribution with 0-mean and $\frac{1}{\sqrt{|\mathcal{V}|}}$-standard deviation:
\begin{equation}
    \hat{\mathbf{Z}}_{\mathcal{V}} = \frac{\mathbf{Z}_{\mathcal{V}}-\mu(\mathbf{Z}_{\mathcal{V}})}{\sigma(\mathbf{Z}_{\mathcal{V}}) \times \sqrt{|\mathcal{V}|}}
\end{equation}
where $\mu(\cdot), \sigma(\cdot)$ denote computing mean value and standard deviation for each feature dimension, respectively. Next, with normalized matrices $\hat{\mathbf{Z}}_{1, \mathcal{V}}$ and $\hat{\mathbf{Z}}_{2, \mathcal{V}}$ for two augmented views, we construct node-level CCA loss $\mathcal{L}_{N}$ which consists of an invariance term $\mathcal{L}_{N}^{inv}$ and a decorrelation term $\mathcal{L}_{N}^{dec}$ as:
\begin{equation}
    \mathcal{L}_{N}^{inv}  = \big|\big|\hat{\mathbf{Z}}_{1,\mathcal{V}}-\hat{\mathbf{Z}}_{2,\mathcal{V}}\big|\big|^{2}_{F}
\end{equation}
\begin{equation}
    \mathcal{L}_{N}^{dec}  = \big|\big| \hat{\mathbf{Z}}_{1,\mathcal{V}}^{T}\hat{\mathbf{Z}}_{1,\mathcal{V}}-\mathbf{I}\big|\big|^{2}_{F}+ \big|\big|\hat{\mathbf{Z}}_{2,\mathcal{V}}^{T}\hat{\mathbf{Z}}_{2,\mathcal{V}}-\mathbf{I}\big|\big|^{2}_{F}
\end{equation}
\begin{equation}
\mathcal{L}_{N} = \mathcal{L}_{N}^{inv} + \lambda_{N}\mathcal{L}_{N}^{dec} 
\end{equation}
where $\mathbf{I} \in \mathbb{R}^{D \times D}$ is the indentity matrix and $\lambda_{N}$ represents a non-negative weight. $\big|\big| \cdot \big|\big|_{F}^{2}$ denote the square of Frobenius norm. $\mathcal{L}_{N}^{inv}$ preserves the node-wise invariant information while $\mathcal{L}_{N}^{dec}$ can prevent dimension collapse~\cite{hua2021feature}, leading to more discriminative representations. 

\paragraph{Group-level CCA objective.}


Group-level SSL signal aims to distinguish the representations of the same hyperedge in the two augmented views from other hyperedge representations, which helps the model preserve group-wise information in the hypergraph. Similar to the node-wise SSL signal, we can formulate the CCA loss in group-wise relation as:
\begin{equation}
    \mathcal{L}_{G}^{inv}  = \big|\big|\hat{\mathbf{Z}}_{1,\mathcal{E}}-\hat{\mathbf{Z}}_{2,\mathcal{E}}\big|\big|^{2}_{F}
\end{equation}
\begin{equation}
    \mathcal{L}_{G}^{dec}  = \big|\big| \hat{\mathbf{Z}}_{1,\mathcal{E}}^{T}\hat{\mathbf{Z}}_{1,\mathcal{E}}-\mathbf{I}\big|\big|^{2}_{F}+ \big|\big|\hat{\mathbf{Z}}_{2,\mathcal{E}}^{T}\hat{\mathbf{Z}}_{2,\mathcal{E}}-\mathbf{I}\big|\big|^{2}_{F}
\end{equation}
\begin{equation}
\mathcal{L}_{G} = \mathcal{L}_{G}^{inv} + \lambda_{G}\mathcal{L}_{G}^{dec} 
\end{equation}
where $\hat{\mathbf{Z}}_{1,\mathcal{E}}$ and $\hat{\mathbf{Z}}_{2,\mathcal{E}}$ are normalized hypergraph embedding matrices from two augmented views. $\lambda_{G}$ is the non-negative coefficient for the decorrelation term. $\mathcal{L}_{G},\mathcal{L}_{G}^{inv},\mathcal{L}_{G}^{dec}$ represent group-wise SSL loss, group-wise invariant term, and group-wise decorrelation term, respectively.

\paragraph{Hierarchical membership-level contrast.}

\begin{figure}[tb!]
\centering
\includegraphics[width=1\linewidth]{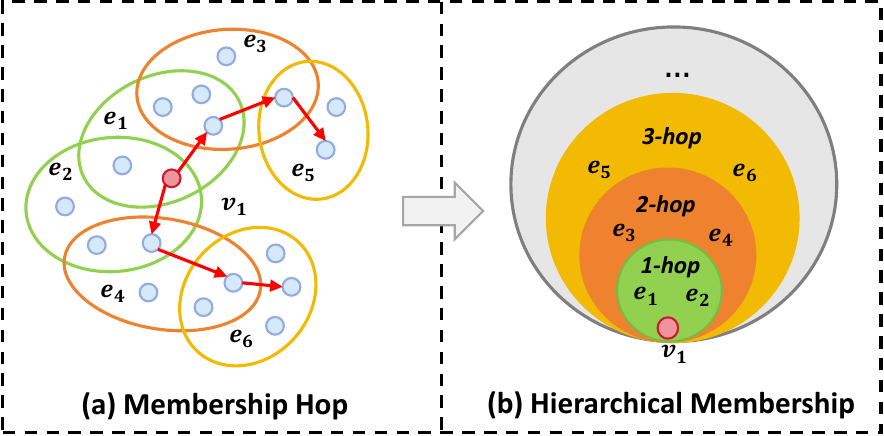}
\caption{Hierachical membership relation}
\label{fig:softmembership}
\end{figure}

Membership-level contrast aims to differentiate a “real” node-hyperedge membership from a “fake” one across the
two augmented views initially. In this work, we only sample and contrast membership pairs from the original view as shown in Figure~\ref{fig:arch}. This can avoid bias caused by random augmentation. Following the instance-level discrimination in the previous work, each anchor node $v_{i}$ and one of its incident hyperedge $e_{j}$ would form a positive pair and all $(v_{i},e_{k})$ where $v_{i} \notin e_{k}$ are considered as negative pairs with the same status. However, this kind of contrast strategy may result in learning bias. For instance, consider the hypergraph in Figure~\ref{fig:softmembership}(a) as a co-author network,
where $v_{1} \in e_{1}$, but $v_{1} \notin e_{3}$. We can observe a close collaboration relationship (significant overlap) between authors $e_{1}$ and $e_{3}$, indicating shared research interests. Although paper $v_{1}$ has a stronger correlation with incident author $e_{1}$, its content may also be similar to works of neighborhood author $e_{3}$. This membership-level similarity strength can be easily extended to a broader neighborhood through cascading overlap, leading to a hierarchical membership structure. Based on this insight, we formally introduce two concepts: membership hop and $k$-hop membership set.

\begin{definition}
\textbf{\textit{(membership hop)}} Given a hypergraph $\mathcal{H}(\mathcal{V},\mathcal{E})$, we connect each vertex with other vertices at the same hyperedge to form a clique expansion graph. For each $v_{i} \in \mathcal{V}$ and $e_{j} \in \mathcal{E}$, the membership hop is defined as:
$$
    \mathcal{M}_{hop}(v_{i},e_{j})=\max\{hop(v_{i},v_{k})|v_{k} \in e_{j} \}
$$
where $hop(v_{i},v_{k})$ denotes the hop number between $v_{i}$ and $v_{k}$ in the expansion graph. In particular, if $e_{j}$ is not reachable from $v_{i}$, $\mathcal{M}_{hop}(v_{i},e_{j})=\infty$.
\end{definition}
\begin{definition}
\textbf{\textit{($k$-hop membership set)}} Given a hypergraph $\mathcal{H}(\mathcal{V},\mathcal{E})$, for each $v_{i} \in \mathcal{V}$, the $v_{i}$'s $k$-hop membership set is:
$$
    \mathcal{M}_{k}(v_{i})=\{e_{j}|\mathcal{M}_{hop}(v_{i},e_{j})=k\}
$$
\end{definition}
This hierarchical membership relation has been illustrated in Figure~\ref{fig:softmembership}. To describe it as a self-supervised signal during training, in $k$-th membership hop, we consider each $\mathcal{M}_{k}(v_{i})$ as the positive set $\mathcal{P}_{k}$ while $\mathcal{M}_{k+1}(v_{i})$ as the negative set $\mathcal{N}_{k}$. Inspired by MIL-NCE~\cite{miech2020end} which can handle multiple positive samples, we design a loss function in $k$-th membership hop for $v_{i}$ as:

\begin{equation}
    \mathcal{L}_{k}(v_{i})=-\mathrm{log}\frac{\sum_{e_{p} \in \mathcal{P}_{k}}e^{\mathcal{D}(v_{i},e_{p}) / \tau}}{\sum_{e_{p} \in \mathcal{P}_{k}}e^{\mathcal{D}(v_{i},e_{p}) / \tau}+\sum_{e_{q} \in \mathcal{N}_{k}}e^{\mathcal{D}(v_{i},e_{q}) / \tau}}
\end{equation}
where $\mathcal{D}(v,e)$ represents a trainable discriminator that predict the similarity score between $v$ and $e$. we implement it using a bilinear function, as described in~\cite{lee2023m}. $\tau$ denotes the temperature parameter. In this way, our objective can be interpreted as hyperedges in $\mathcal{M}_{k}(v_{i})$ is overall more similar with $v_{i}$ than those in $\mathcal{M}_{k+1}(v_{i})$. This multi-positive setting can enhance model robustness to noisy positive samples. Then, it is natural to enumerate this objective in different membership ranges, formulating our final hierarchical membership loss $\mathcal{L}_{HM}$ as:
\begin{equation}
    \mathcal{L}_{HM}=-\sum_{v_{i} \in \mathcal{V}}\frac{1}{K}\sum_{k=1}^{K}\mathrm{log}\Big[\min\{e^{-\mathcal{L}_{k}(v_{i})},\alpha\}\Big]
\end{equation}
where $\alpha \in (0,1)$ is the secure threshold to avoid over-pushing two neighboring membership hops. $K$ is a hyperparameter that denotes the chosen membership range. However, using all the samples in different membership sets is computationally prohibitive for large hypergraphs. To deal with this, we uniformly sample a fixed number of $d$ hyperedges in each $\mathcal{M}_{k}(v_{i})$ in implementation. The experiments in Section 4 show that only a small sample number can make SE-HSSL achieve superior performance than the state-of-the-art.

 We consider the frequency of calculating similarity function $D(v,e)$ when evaluating the time complexity in membership contrast. The current state-of-the-art method TriCL requires a complexity of up to $O(|\mathcal{V}| \times |\mathcal{E}|)$, whereas the time complexity of SE-HSSL is $O(|\mathcal{V}|Kd)$.

\subsection{Model Optimization}

To preserve multi-view structural information during self-supervised hypergraph representation learning, we combine tri-directional signals and jointly optimize the node-level CCA loss $\mathcal{L}_{N}$, group-level CCA loss $\mathcal{L}_{G}$, and hierarchical membership-level contrast loss $\mathcal{L}_{HM}$ as: 
\begin{equation}
    \mathcal{L} = \mathcal{L}_{N}+\lambda_{1}\mathcal{L}_{G}+\lambda_{2}\mathcal{L}_{HM}+\lambda_{3}||\mathbf{\Theta}||^{2}
\end{equation}
where $\lambda_{1},\lambda_{2},\lambda_{3}$ are hyperparameters to control the strengths of group-level loss, membership-level loss, and $L_{2}$ regularization respectively. $\mathbf{\Theta}=\{\mathbf{\Theta}_{\mathcal{E}}, \mathbf{\Theta}_{\mathcal{V}}\}$ represents the learnable parameters in hypergraph encoder.

\section{Experiment}

In this section, We first briefly introduce the experimental setups. Subsequently, we conduct extensive experiments to validate the effectiveness and efficiency of our model.


\subsection{Experimental Setups}

\paragraph{Datasets.} We empirically evaluate the model performance on 7 commonly used hypergraph benchmark datasets, whose details are shown in Table~\ref{tab:stat}.

\paragraph{Baselines.} To comprehensively validate our method, we compare it against 11 strong baselines, including 4 semi-supervised models (i.e., HGNN~\cite{Yifan:19}, HyperGCN~\cite{Yadati:19}, UniGIN~\cite{cai2022hypergraph}), and AllSet~\cite{chien2021you}, 5 common graph based self-supervised methods (i.e., DGI~\cite{velivckovic2018deep}, GRACE~\cite{zhu2020deep}, BGRL~\cite{thakoor2021bootstrapped}, CCA-SSG~\cite{zhang2021canonical}, and COSTA~\cite{zhang2022costa}, and 2 hypergraph contrastive learning methods (i.e., HyperGCL~\cite{wei2022augmentations} and the current SOTA method TriCL~\cite{lee2023m}). For the evaluation of graph-based SSL methods, we follow~\cite{lee2023m} by applying clique expansion to hypergraphs, thereby converting them into simple graphs.

\paragraph{Evaluation protocol.} In this work, we focus on two commonly used benchmark hypergraph learning tasks: node classification and node clustering~\cite{zhou2006learning,lee2023m}. For the node classification task, we follow the linear evaluation scheme utilized in previous work~\cite{wei2022augmentations,lee2023m}. Specifically, we first train the hypergraph encoder in a self-supervised manner as described in Section 3. Afterward, with the trained model, we generate node embeddings and randomly split them into training, validation, and test samples using splitting percentages of 10\%, 10\%, and 80\%, respectively. Finally, we employ the obtained embeddings from the training set to train a logistic regression classifier and evaluate its performance on test node embeddings. For fair evaluation, we report the average test accuracy along with its corresponding standard deviation based on 20 random initializations for each dataset. 

For the node clustering task~\cite{lee2023m}, we apply the k-means clustering algorithm to the output node representations and get the predicted results. As in~\cite{zhao2021graph}, the normalized mutual information (NMI), and adjusted rand index (ARI) are used to measure the performance of clustering. We randomly perform k-means on generated node embeddings 5 times and report the averaged results.

\begin{table}[t!]
\caption{Statistics of datasets}
\vspace{-5px}
\centering
\tabcolsep 0.3pt
\renewcommand\arraystretch{1.05}
\footnotesize
\begin{tabular}{cccccc}
\toprule
Name & Type & $|\mathcal{V}|$ & $|\mathcal{E}|$ & $F$ & Classes \\ \midrule
Cora~\cite{wei2022augmentations} & co-citation & 2708   &  1579     & 1433  &  7      \\
Citeseer~\cite{wei2022augmentations} & co-citation & 3312    & 1079       &  3703 & 6       \\ 
Pubmed~\cite{wei2022augmentations}  & co-citation & 19717    & 7963  & 500 &  3       \\ 
Cora-CA~\cite{Yadati:19} & co-author &  2708   &  1072   &  1433    & 7   \\ 
NTU2012~\cite{yang2022semi}  & graphics &  2012   &   2012 & 100  & 67       \\ 
ModelNet40~\cite{yang2022semi} & graphics &  12311   &  12311   & 100 & 40         \\
Zoo~\cite{hein2013total}  & animal &  101   &  43 &  16  & 7        \\\bottomrule 
\end{tabular}
\label{tab:stat}
\end{table}

\paragraph{Implementation details.} For all baselines,
we report their performance based on the official implementations and default hyper-parameters from original papers. For SE-HSSL, we set the number of encoder layers to 1 for all datasets. We set the group-level coefficient $\lambda_{1}=1$ and regularization coefficient $\lambda_{3}=0.05$. The sampling number $d$ in each membership set is fixed to 10. The membership-level coefficient $\lambda_{2}$ is tuned over $\{0.1,0.18,0.20\}$ . Besides, we tune the embedding size $D$ in $\{256,512,784,1024\}$, the hyperedge hop $K$ in $\{1,2,3,4\}$, the temperature $\tau$ in $\{0.4,0.5,0.6\}$, and threshold $\alpha$ in $\{0.62,0.65\}$. We have released our code in \url{https://github.com/Coco-Hut/SE-HSSL}.



\subsection{Node Classification Evaluation}

\begin{table*}[t]
\caption{Evaluation results for node classification: mean accuracy (\%) ± standard deviation. For each dataset, the top and runner-up performances are indicated by boldface and underlined formatting, respectively.}
\label{tab:classification}
\vspace{-4px}
\centering
\renewcommand\arraystretch{1.05}
\footnotesize
\resizebox{\textwidth}{!}{%
\begin{tabular}{ccccccccc}
\toprule
Method   & Cora & Citeseer & Pubmed & Cora-CA & NTU2012 &  Zoo & ModelNet40 & Avg.Rank \\ \midrule
HGNN     &  73.36$\pm$3.40  &   64.66$\pm$2.95      &  79.22$\pm$1.59    &  73.26$\pm$2.86    &  70.53$\pm$7.68       &    77.56$\pm$11.34       & 91.14$\pm$0.35    &  9.3    \\
HyperGCN     & 74.32$\pm$2.39   &    65.28$\pm$2.85     &  77.90$\pm$1.39    &   72.17$\pm$1.92   &   71.68$\pm$2.40         &   78.37$\pm$10.90    & 90.81$\pm$0.30    &  8.7    \\
UniGIN     &  71.81$\pm$2.36  &   63.58$\pm$3.07      &   77.71$\pm$3.39   &  71.20$\pm$2.50    &    68.83$\pm$3.60         &   74.51$\pm$10.87    &  90.65$\pm$0.71   &  11.0    \\
AllSet     & 77.12$\pm$0.61    &   67.88$\pm$1.36      &   81.55$\pm$0.29    &    76.89$\pm$1.63            &   73.70$\pm$0.50       &   78.19$\pm$10.39  &   95.92$\pm$0.24        &  5.4        \\ \midrule
DGI     & 76.54$\pm$1.48   &   66.78$\pm$1.56      &  77.68$\pm$2.24    & 74.56$\pm$1.44     &   70.46$\pm$2.45         &   63.11$\pm$13.14    &  91.48$\pm$0.57   &  9.7    \\
GRACE  & 76.90$\pm$2.25 & 66.45$\pm$2.57 & 80.64$\pm$0.46 & 75.02$\pm$2.06 & 69.77$\pm$2.38 & 62.16$\pm$15.45 & 89.96$\pm$0.39 & 9.7  \\
BGRL &  78.01$\pm$1.53  &  67.11$\pm$1.78  &  81.38$\pm$0.61    &  77.97$\pm$2.84    &    71.49$\pm$1.78     &     65.50$\pm$12.36      &   92.28$\pm$2.52  &  6.9    \\
CCA-SSG    & 79.29$\pm$1.51   &   67.48$\pm$1.77    &  82.14$\pm$0.64    & 78.45$\pm$1.16    &   72.81$\pm$2.43   &   78.35$\pm$10.85    &  93.91$\pm$0.18   &  4.3    \\
COSTA     &  77.49$\pm$1.83  &   68.74$\pm$1.61   &  81.89$\pm$0.56    &   78.03$\pm$2.07   &     72.14$\pm$2.18     &   69.69$\pm$9.46    &   91.64$\pm$2.18  &  5.7    \\ \midrule
HyperGCL     & 78.86$\pm$1.09   &    69.49$\pm$1.62     &  83.95$\pm$0.40  &   77.90$\pm$1.15   &          \textbf{75.44$\pm$1.15}   &  67.74$\pm$10.34  &   96.42$\pm$0.31  &  4.1    \\
TriCL    & \underline{81.61$\pm$1.24}   &    \underline{72.05$\pm$1.14}     &  \underline{84.24$\pm$0.63}    &   \underline{82.17$\pm$0.92}   &     \underline{75.21$\pm$2.46}        &   \underline{80.09$\pm$11.14}    &  \underline{97.05$\pm$0.13}   &  \underline{2.0}    \\
SE-HSSL     &  \textbf{82.47$\pm$1.14}   &    \textbf{72.74$\pm$0.98}     &   \textbf{84.38$\pm$0.71}     &    \textbf{82.59$\pm$0.93}      &   75.19$\pm$2.36         &   \textbf{80.68$\pm$11.13}  &    \textbf{97.17$\pm$0.13}        &    \textbf{1.3}      \\ \bottomrule
\end{tabular}}
\vspace{-8pt}
\end{table*}

We first employ the node classification task to evaluate the effectiveness of SE-HSSL. The performance of all compared methods is summarized in Table~\ref{tab:classification}. The first 4 lines present the results of supervised HGNNs. It can be observed that the AllSet network demonstrates superior performance compared to other methods. This is because its learnable multiset functions allow for higher expressive power. However, we notice that these methods perform poorly compared to the HSSL models. This is because, when labels are limited, supervised models are prone to overfitting and fail to generalize well to unseen data. For SSL frameworks, they can capture the underlying semantics of data without labels. Lines 5-9 show the results of several popular graph SSL methods. Although they achieve superior overall performance compared to supervised methods, their accuracy scores are considerably lower than those of HSSL methods. This is because, during the expansion of a hypergraph to a simple graph, a significant amount of high-order information is lost. The above observations suggest that it is necessary to devise specific SSL models for hypergraph learning. The last 3 lines compare the results of SE-HSSL with the latest HCL models. The proposed SE-HSSL achieves the best performance on most datasets. For example, we achieve mean accuracy scores of 82.47\% and 72.74\% on the Cora and Citeseer datasets, respectively. These results show a relative improvement of 0.86\% and 0.69\% over the current SOAT method TriCL. It is noteworthy that SE-HSSL outperforms TriCL and HyperGCL even without the utilization of node-wise and group-wise negative samples. Furthermore, it only requires a small number of membership pairs. These highlight the effectiveness of our tri-directional sampling-efficient signals.

\subsection{Node Clustering Evaluation}

\begin{table*}[t]
\caption{Evaluation results for node clustering: NMI and ARI}
\label{tab:clustering}
\vspace{-4px}
\centering
\renewcommand\arraystretch{1.05}
\resizebox{\textwidth}{!}{%
\begin{tabular}{c cc cc cc cc cc cc cc c}
\toprule
\multirow{2}{*}{Method} & \multicolumn{2}{c}{Cora} & \multicolumn{2}{c}{Citeseer} & \multicolumn{2}{c}{Pubmed} & \multicolumn{2}{c}{Cora-CA} & \multicolumn{2}{c}{NTU2012}  & \multicolumn{2}{c}{Zoo} & \multicolumn{2}{c}{ModelNet40} & \multirow{2}{*}{Avg.Rank} \\ \cline{2-15} 
                        &    NMI  $\uparrow$   &     ARI $\uparrow$    &     NMI $\uparrow$       &    ARI $\uparrow$  &     NMI $\uparrow$       &    ARI $\uparrow$ &     NMI  $\uparrow$       &    ARI $\uparrow$ &     NMI $\uparrow$       &    ARI $\uparrow$ &       NMI   $\uparrow$     &    ARI $\uparrow$ &     NMI  $\uparrow$       &    ARI $\uparrow$  \\ \midrule
BGRL       &     40.81       &     22.54       &   31.69       & 24.87  &       15.98        &  16.80 &       32.48        &  21.72 &      67.76         & 33.52 &     70.55          &58.65 &      73.63         &  45.36  & 5.9 \\
CCA-SSG        & 49.08            &    40.57         &   38.45         &     36.02                         & 17.22  &        17.24       & 38.13 &       26.18        & 73.72  &      42.26         & 79.59 &      74.70         & 79.08 &     56.44           & 3.9 \\
COSTA                    &     45.58        &    36.08        &    34.83   &  28.18  &      20.41      &  15.92  &   36.09            &  22.90   &     71.77          &   40.32  &    76.25     &  61.95  &    75.23  & 46.33   & 4.9 \\
HyperGCL   &  40.24  &    32.35    &    39.06    &  38.88  &  28.72             &  24.73  &      \underline{46.15}   &  \underline{39.33} &   \textbf{84.27}  & \textbf{70.88}  & 77.40              & 79.36 &   93.71  &  \underline{89.91}  & 2.9 \\
TriCL                    &      \underline{56.95}       &    \underline{49.77}     &     \underline{42.63}  &    \underline{42.44}   &     \underline{30.80}    &  \underline{29.06}  & 44.39             &  37.21 &     \underline{83.75}          & \underline{68.33}  &   \underline{90.87}    &  \underline{88.62} &   \underline{93.80} &  87.07  & \underline{2.2} \\
SE-HSSL    &    \textbf{59.41}   &   \textbf{53.37}  &  \textbf{43.66}  &  \textbf{43.01}  &    \textbf{32.79}          & \textbf{29.73}  &      \textbf{50.56}     &  \textbf{44.47} &   83.31 &  68.26  &  \textbf{91.71}   & \textbf{91.48} & \textbf{94.36}                           &  \textbf{91.63} & \textbf{1.3}\\
\bottomrule
\end{tabular}}
\vspace{-6px}
\end{table*}

In the subsection, we conduct the node clustering task on 7 datasets. We compare SE-HSSL with 2 HCL methods and 3 strong graph SSL models. Table~\ref{tab:clustering} shows that SE-HSSL achieves the best clustering results on 6 out of 7 datasets and obtains 1st place in terms of the average rank. Moreover, we find it achieves an average improvement of 1.80\% in NMI and 2.78\% in ARI compared to TriCL across all datasets. This is because our sampling-efficient signals effectively reduce training bias and avoid degenerated solutions, making learned embeddings more informative and discriminative.

\subsection{Ablation Study}

\begin{table*}[t]
\caption{Ablation study on different SSL signals. The variants SE-HSSL-w/o N, SE-HSSL-w/o G, and SE-HSSL-w/o HM represent models that have removed node-level CCA loss, group-level CCA loss, and hierarchical membership loss, respectively.}
\vspace{-4px}
\centering
\renewcommand\arraystretch{1.05}
\resizebox{\textwidth}{!}{%
\begin{tabular}{ccccccccc}
\toprule
Method   & Cora & Citeseer & Pubmed & Cora-CA & NTU2012 &  Zoo & ModelNet40 & Avg.Rank \\ \midrule
SE-HSSL-w/o N     & 80.21$\pm$1.51  &  72.16$\pm$1.14      & 83.09$\pm0.71$ &  80.04$\pm$2.34  & 74.22$\pm$2.68 & \underline{80.31$\pm$10.94}  &  97.13$\pm$0.10  & 3.4     \\
SE-HSSL-w/o G     & 81.23$\pm$0.85  &  \underline{72.28$\pm$0.95}      & 83.84$\pm$0.74 &  \underline{82.27$\pm$0.93}  & \underline{75.08$\pm$2.21}   & 79.81$\pm$10.88  & 97.12$\pm$0.11   & \underline{2.7}     \\
SE-HSSL-w/o HM     & \underline{81.71$\pm$1.41}  &  71.67$\pm$1.21      & \underline{83.90$\pm$0.76} &  82.15$\pm$1.02  & 74.81$\pm$2.25   & 79.51$\pm$10.89  & \underline{97.15$\pm$0.14}   &   2.9   \\
SE-HSSL     &  \textbf{82.47$\pm$1.14}   &    \textbf{72.74$\pm$0.98}     &   \textbf{84.38$\pm$0.71}     &    \textbf{82.59$\pm$0.93}      &   \textbf{75.19$\pm$2.36}         &   \textbf{80.68$\pm$11.13}  &    \textbf{97.17$\pm$0.13}        &    \textbf{1.0}      \\ \bottomrule
\end{tabular}}
\label{tab:ablation}
\vspace{-6pt}
\end{table*}

In this subsection, we evaluate the effectiveness of each designed SSL objective. The results are summarized
in Table~\ref{tab:ablation}. We find that the performance degrades when any of the objectives is dropped, which indicates the effectiveness of each signal. We also observe that SE-HSSL-w/o N achieves lower accuracy compared to the other two variants on most datasets. This is because we focus on the node-level downstream task. Notably, the removal of hierarchical membership loss brings about even larger performance degradation than SE-HSSL-w/o N on Citeseer and Zoo datasets, which shows that this signal is essential in boosting model performance.

\subsection{Parameter Sensitivity Analysis}

\begin{figure}[th]
	\centering
	\subfloat[Embedding size $D$]{
		\includegraphics[width=0.5\linewidth]{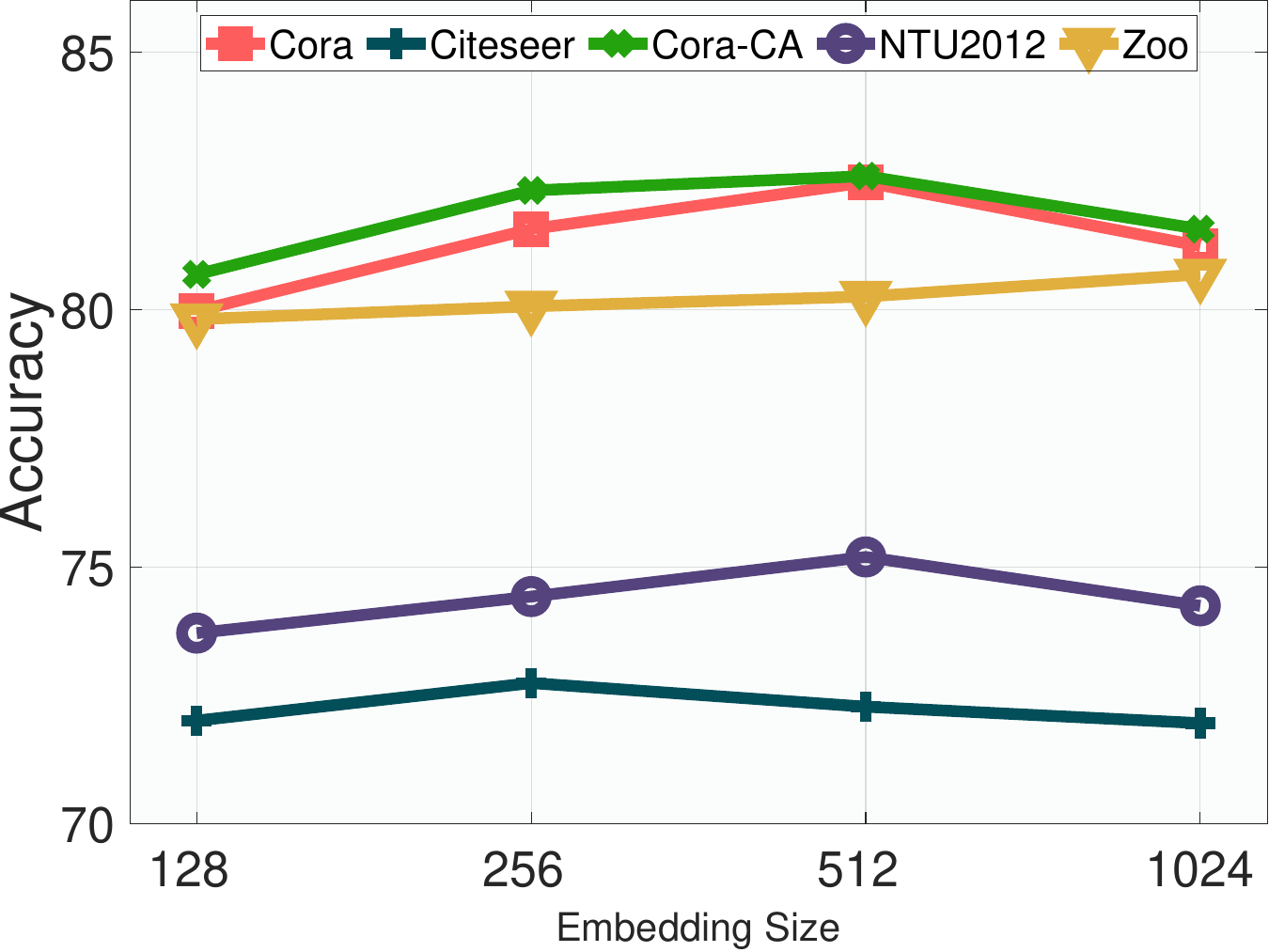}}
	\subfloat[Hyperedge hop $K$]{
		\includegraphics[width=0.5\linewidth]{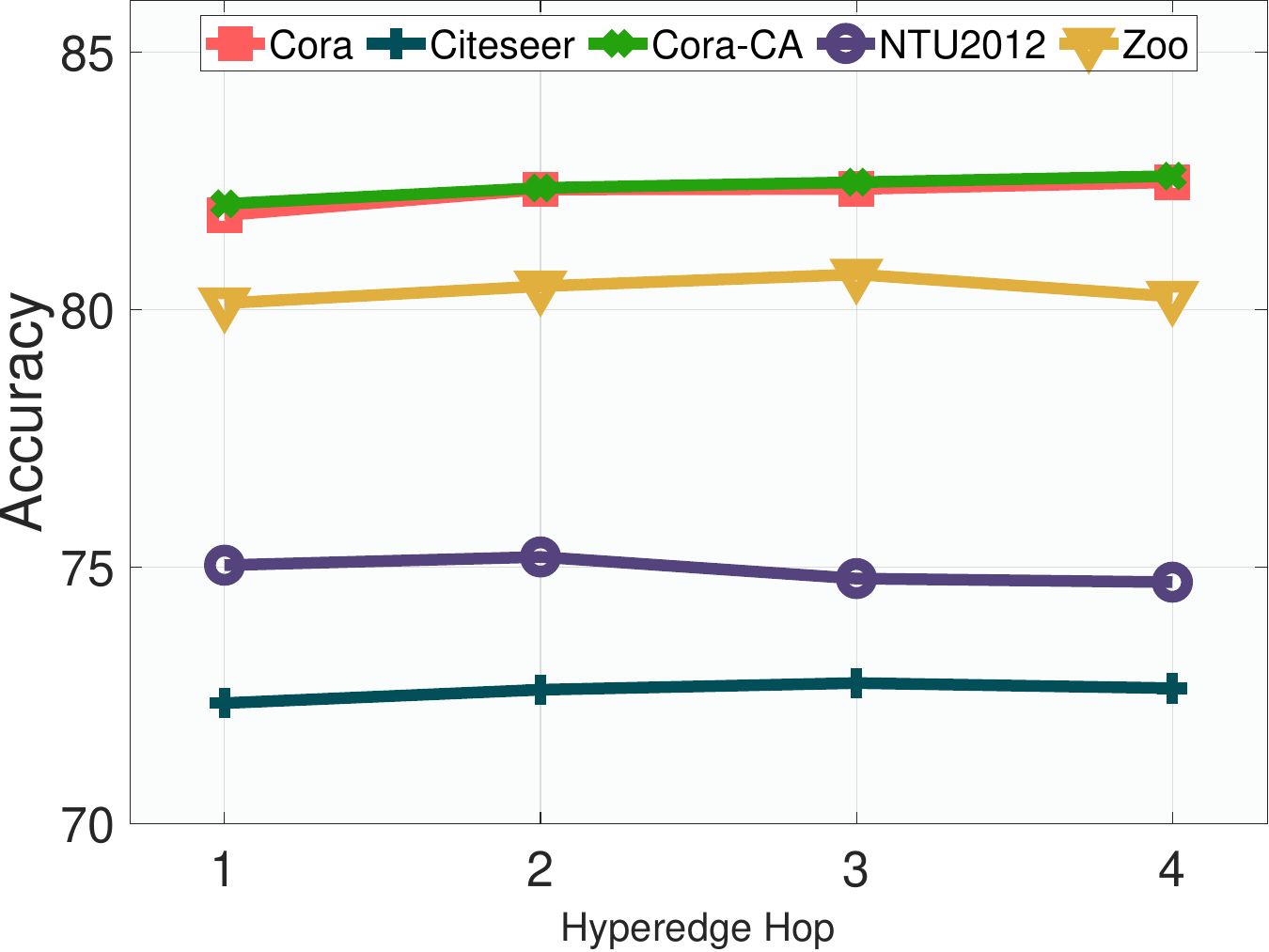}}
        \caption{Parameter sensitivity test.}
        \label{fig:params}
\end{figure}

We perform sensitivity analysis on two critical parameters to further validate the robustness of SE-HSSL. We report mean accuracy w.r.t. the node classification
task in Figure~\ref{fig:params}.

\paragraph{Embedding size $D$.} Figure~\ref{fig:params}(a) illustrates that as we vary the embedding size from 128 to 512, the model performance improves overall, suggesting that our model benefits from a larger representation capacity. However, we notice a
decline in performance as the embedding size further increases on most datasets. This is because the embedding space becomes sparse when $D$ is too large. Consequently, there is limited meaningful information available in each dimension of the embedding, with many dimensions being irrelevant or redundant. This makes it challenging for the downstream classifier to extract meaningful features from embeddings.

\paragraph{Hyperedge hop $K$.} In the hierarchical membership contrast, we sample neighborhood hyperedges ranging from 1 to $K$ hops for each node. As depicted in Figure~\ref{fig:params}(b), we observe that the model can achieve near-optimal performance when $K$ is set to 1. Increasing the neighborhood range only leads to marginal performance gains on most datasets. This suggests that our model exhibits robustness and is not highly sensitive to the number of hops when sampling hyperedges. We attribute this to our multi-positive membership loss which does not enforce a higher similarity for all positive
pairs and can effectively disregard the noisy positives.

\subsection{Efficiency Evaluation}

\begin{figure}[t]
\centering
\includegraphics[width=0.75\linewidth]{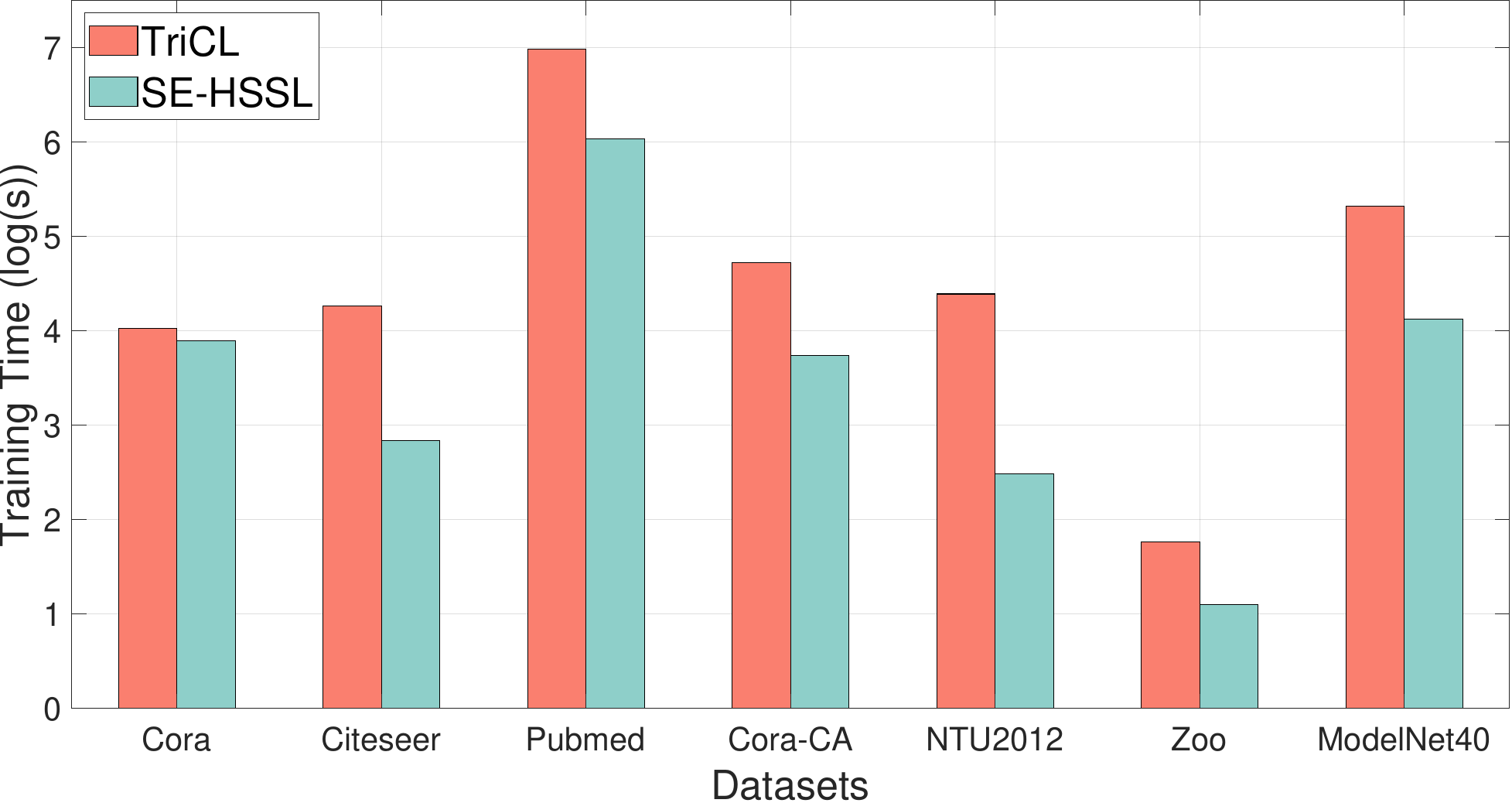}
\caption{The training time comparison between SE-HSSL and TriCL. log($\cdot$) represents the natural logarithm.}
\label{fig:efficiency}
\end{figure}

We measure the training time to validate the efficiency of the proposed SE-HSSL. We mainly compare it with TriCL, which is the current SOTA approach with tri-directional contrast objectives. The results are shown in Figure~\ref{fig:efficiency}. For better visibility, we take the logarithm of training time when plotting. It can be found that SE-HSSL is consistently faster than TriCL across all datasets. What stands out is that our model is at least 2.5 times faster than TriCL on large datasets, such as PubMed and ModelNet40. This efficiency improvement can be attributed to three key factors: 1) Sampling-free CCA objectives in node-level and group-level learning reduce computation burden during training. 2) Hierarchical contrast in membership-level learning only requires a small number of node-hyperedge pairs in each neighborhood range, thereby considerably enhancing the training speed. 3) Our method is projection head-free, further reducing the computation cost.

\section{Related Work}

\subsection{Contrastive Learning on Graph}

Contrastive learning (CL), which was originally utilized in computer vision~\cite{he2020momentum}, has recently been applied to the graph domain. The basic idea of graph contrastive learning (GCL) is to maximize agreement between instances (e.g., node, subgraph, and graph) of
different views augmented from the original graph. DGI~\cite{velivckovic2018deep} maximizes the mutual information between the local and global representations of nodes to learn their embeddings. GRACE~\cite{zhu2020deep} focuses on contrasting views at the node level. GraphCL~\cite{you2020graph} and GCA~\cite{zhu2021graph} employ multiple augmentation strategies and treat nodes/graphs with distinct IDs from the anchor node/graph as negative samples. BGRL and CCA-SSG~\cite{zhang2021canonical} are two methods that alleviate the need for contrasting with tons of negative samples, thereby improving efficiency and reducing training bias. COSTA~\cite{zhang2022costa} and SFA~\cite{zhang2023spectral} are two feature-level augmentation strategies that aim to construct more effective contrast views than random augmentations. However, the above methods can not effectively learn high-order information in hypergraphs, so it is necessary to design specific HSSL approaches.


\subsection{Contrastive Learning on Hypergraph}
Hypergraph contrastive learning has received little attention and remains largely unexplored. While some investigations in recommender systems have applied contrastive strategy in hypergraphs~\cite {yang2021hyper,xia2022hypergraph}, they are not designed for general HCL. HyperGCL~\cite{wei2022augmentations} is the first general HCL framework. It proposes a variational hypergraph auto-encoder to generate augmentation views that effectively preserve the high-order information in the original hypergraph. TriCL~\cite{lee2023m} utilizes tri-directional contrastive objectives to further capture group-level and membership-level information in hypergraphs. However, existing methods select a large number of negative samples arbitrarily, which can be both unreliable and inefficient. In this work, we explore approaches to construct informative and reliable hypergraph SSL signals without the need for an abundance of negative samples.

\section{Conclusion}

In this paper, we introduce SE-HSSL, a sampling-efficient hypergraph self-supervised learning framework, to address the efficiency bottleneck and training bias in instance-level contrastive methods. To achieve this, we propose sampling-free CCA-based objectives as self-supervised signals for learning at both the node and group levels. This approach allows us to effectively avoid degenerate solutions in the learned embeddings and reduce sampling bias. Additionally, we develop a novel hierarchical membership contrast objective, which only requires a small number of node-hyperedge pairs to achieve effective membership-level self-supervised training. Through extensive experiments conducted on 7 real-world datasets, we empirically demonstrate the superior effectiveness and efficiency of SE-HSSL.

\appendix

\bibliographystyle{named}
\bibliography{ijcai24}

\begin{thebibliography}{}

\bibitem[\protect\citeauthoryear{Andrew \bgroup \em et al.\egroup }{2013}]{andrew2013deep}
Galen Andrew, Raman Arora, Jeff Bilmes, and Karen Livescu.
\newblock Deep canonical correlation analysis.
\newblock In {\em International conference on machine learning}, pages 1247--1255. PMLR, 2013.

\bibitem[\protect\citeauthoryear{Antelmi \bgroup \em et al.\egroup }{2023}]{antelmi2023survey}
Alessia Antelmi, Gennaro Cordasco, Mirko Polato, Vittorio Scarano, Carmine Spagnuolo, and Dingqi Yang.
\newblock A survey on hypergraph representation learning.
\newblock {\em ACM Computing Surveys}, 56(1):1--38, 2023.

\bibitem[\protect\citeauthoryear{Cai \bgroup \em et al.\egroup }{2022}]{cai2022hypergraph}
Derun Cai, Moxian Song, Chenxi Sun, Baofeng Zhang, Shenda Hong, and Hongyan Li.
\newblock Hypergraph structure learning for hypergraph neural networks.
\newblock In {\em Proceedings of the Thirty-First International Joint Conference on Artificial Intelligence, IJCAI-22}, pages 1923--1929, 2022.

\bibitem[\protect\citeauthoryear{Chang \bgroup \em et al.\egroup }{2018}]{chang2018scalable}
Xiaobin Chang, Tao Xiang, and Timothy~M Hospedales.
\newblock Scalable and effective deep cca via soft decorrelation.
\newblock In {\em Proceedings of the IEEE Conference on Computer Vision and Pattern Recognition}, pages 1488--1497, 2018.

\bibitem[\protect\citeauthoryear{Chien \bgroup \em et al.\egroup }{2021}]{chien2021you}
Eli Chien, Chao Pan, Jianhao Peng, and Olgica Milenkovic.
\newblock You are allset: A multiset function framework for hypergraph neural networks.
\newblock In {\em International Conference on Learning Representations}, 2021.

\bibitem[\protect\citeauthoryear{Feng \bgroup \em et al.\egroup }{2019}]{Yifan:19}
Yifan Feng, Haoxuan You, Zizhao Zhang, Rongrong Ji, and Yue Gao.
\newblock Hypergraph neural networks.
\newblock In {\em Proceedings of the AAAI conference on artificial intelligence}, volume~33, pages 3558--3565, 2019.

\bibitem[\protect\citeauthoryear{He \bgroup \em et al.\egroup }{2020}]{he2020momentum}
Kaiming He, Haoqi Fan, Yuxin Wu, Saining Xie, and Ross Girshick.
\newblock Momentum contrast for unsupervised visual representation learning.
\newblock In {\em Proceedings of the IEEE/CVF conference on computer vision and pattern recognition}, pages 9729--9738, 2020.

\bibitem[\protect\citeauthoryear{Hein \bgroup \em et al.\egroup }{2013}]{hein2013total}
Matthias Hein, Simon Setzer, Leonardo Jost, and Syama~Sundar Rangapuram.
\newblock The total variation on hypergraphs-learning on hypergraphs revisited.
\newblock {\em Advances in Neural Information Processing Systems}, 26, 2013.

\bibitem[\protect\citeauthoryear{Hua \bgroup \em et al.\egroup }{2021}]{hua2021feature}
Tianyu Hua, Wenxiao Wang, Zihui Xue, Sucheng Ren, Yue Wang, and Hang Zhao.
\newblock On feature decorrelation in self-supervised learning.
\newblock In {\em Proceedings of the IEEE/CVF International Conference on Computer Vision}, pages 9598--9608, 2021.

\bibitem[\protect\citeauthoryear{Jia \bgroup \em et al.\egroup }{2021}]{jia2021hypergraph}
Renqi Jia, Xiaofei Zhou, Linhua Dong, and Shirui Pan.
\newblock Hypergraph convolutional network for group recommendation.
\newblock In {\em 2021 IEEE International Conference on Data Mining (ICDM)}, pages 260--269. IEEE, 2021.

\bibitem[\protect\citeauthoryear{Jing \bgroup \em et al.\egroup }{2014}]{jing2014intra}
Xiao-Yuan Jing, Rui-Min Hu, Yang-Ping Zhu, Shan-Shan Wu, Chao Liang, and Jing-Yu Yang.
\newblock Intra-view and inter-view supervised correlation analysis for multi-view feature learning.
\newblock In {\em Proceedings of the AAAI Conference on Artificial Intelligence}, volume~28, 2014.

\bibitem[\protect\citeauthoryear{Kim \bgroup \em et al.\egroup }{2020}]{kim2020hypergraph}
Eun-Sol Kim, Woo~Young Kang, Kyoung-Woon On, Yu-Jung Heo, and Byoung-Tak Zhang.
\newblock Hypergraph attention networks for multimodal learning.
\newblock In {\em Proceedings of the IEEE/CVF conference on computer vision and pattern recognition}, pages 14581--14590, 2020.

\bibitem[\protect\citeauthoryear{Lee and Shin}{2023}]{lee2023m}
Dongjin Lee and Kijung Shin.
\newblock I’m me, we’re us, and i’m us: Tri-directional contrastive learning on hypergraphs.
\newblock In {\em Proceedings of the AAAI Conference on Artificial Intelligence}, volume~37, pages 8456--8464, 2023.

\bibitem[\protect\citeauthoryear{Miech \bgroup \em et al.\egroup }{2020}]{miech2020end}
Antoine Miech, Jean-Baptiste Alayrac, Lucas Smaira, Ivan Laptev, Josef Sivic, and Andrew Zisserman.
\newblock End-to-end learning of visual representations from uncurated instructional videos.
\newblock In {\em Proceedings of the IEEE/CVF Conference on Computer Vision and Pattern Recognition}, pages 9879--9889, 2020.

\bibitem[\protect\citeauthoryear{Thakoor \bgroup \em et al.\egroup }{2021}]{thakoor2021bootstrapped}
Shantanu Thakoor, Corentin Tallec, Mohammad~Gheshlaghi Azar, R{\'e}mi Munos, Petar Veli{\v{c}}kovi{\'c}, and Michal Valko.
\newblock Bootstrapped representation learning on graphs.
\newblock In {\em ICLR 2021 Workshop on Geometrical and Topological Representation Learning}, 2021.

\bibitem[\protect\citeauthoryear{Veli{\v{c}}kovi{\'c} \bgroup \em et al.\egroup }{2018}]{velivckovic2018deep}
Petar Veli{\v{c}}kovi{\'c}, William Fedus, William~L Hamilton, Pietro Li{\`o}, Yoshua Bengio, and R~Devon Hjelm.
\newblock Deep graph infomax.
\newblock In {\em International Conference on Learning Representations}, 2018.

\bibitem[\protect\citeauthoryear{Wei \bgroup \em et al.\egroup }{2022}]{wei2022augmentations}
Tianxin Wei, Yuning You, Tianlong Chen, Yang Shen, Jingrui He, and Zhangyang Wang.
\newblock Augmentations in hypergraph contrastive learning: Fabricated and generative.
\newblock {\em Advances in neural information processing systems}, 35:1909--1922, 2022.

\bibitem[\protect\citeauthoryear{Xia \bgroup \em et al.\egroup }{2022a}]{xia2022hypergraph}
Lianghao Xia, Chao Huang, Yong Xu, Jiashu Zhao, Dawei Yin, and Jimmy Huang.
\newblock Hypergraph contrastive collaborative filtering.
\newblock In {\em Proceedings of the 45th International ACM SIGIR conference on research and development in information retrieval}, pages 70--79, 2022.

\bibitem[\protect\citeauthoryear{Xia \bgroup \em et al.\egroup }{2022b}]{xia2022self}
Lianghao Xia, Chao Huang, and Chuxu Zhang.
\newblock Self-supervised hypergraph transformer for recommender systems.
\newblock In {\em Proceedings of the 28th ACM SIGKDD Conference on Knowledge Discovery and Data Mining}, pages 2100--2109, 2022.

\bibitem[\protect\citeauthoryear{Xiao \bgroup \em et al.\egroup }{2019}]{xiao2019multi}
Li~Xiao, Junqi Wang, Peyman~H Kassani, Yipu Zhang, Yuntong Bai, Julia~M Stephen, Tony~W Wilson, Vince~D Calhoun, and Yu-Ping Wang.
\newblock Multi-hypergraph learning-based brain functional connectivity analysis in fmri data.
\newblock {\em IEEE transactions on medical imaging}, 39(5):1746--1758, 2019.

\bibitem[\protect\citeauthoryear{Yadati \bgroup \em et al.\egroup }{2019}]{Yadati:19}
Naganand Yadati, Madhav Nimishakavi, Prateek Yadav, Vikram Nitin, Anand Louis, and Partha Talukdar.
\newblock Hypergcn: A new method for training graph convolutional networks on hypergraphs.
\newblock {\em Advances in neural information processing systems}, 32, 2019.

\bibitem[\protect\citeauthoryear{Yang \bgroup \em et al.\egroup }{2021}]{yang2021hyper}
Haoran Yang, Hongxu Chen, Lin Li, S~Yu Philip, and Guandong Xu.
\newblock Hyper meta-path contrastive learning for multi-behavior recommendation.
\newblock In {\em 2021 IEEE International Conference on Data Mining (ICDM)}, pages 787--796. IEEE, 2021.

\bibitem[\protect\citeauthoryear{Yang \bgroup \em et al.\egroup }{2022}]{yang2022semi}
Chaoqi Yang, Ruijie Wang, Shuochao Yao, and Tarek Abdelzaher.
\newblock Semi-supervised hypergraph node classification on hypergraph line expansion.
\newblock In {\em Proceedings of the 31st ACM International Conference on Information \& Knowledge Management}, pages 2352--2361, 2022.

\bibitem[\protect\citeauthoryear{Yi and Park}{2020}]{yi2020hypergraph}
Jaehyuk Yi and Jinkyoo Park.
\newblock Hypergraph convolutional recurrent neural network.
\newblock In {\em Proceedings of the 26th ACM SIGKDD international conference on knowledge discovery \& data mining}, pages 3366--3376, 2020.

\bibitem[\protect\citeauthoryear{You \bgroup \em et al.\egroup }{2020}]{you2020graph}
Yuning You, Tianlong Chen, Yongduo Sui, Ting Chen, Zhangyang Wang, and Yang Shen.
\newblock Graph contrastive learning with augmentations.
\newblock {\em Advances in neural information processing systems}, 33:5812--5823, 2020.

\bibitem[\protect\citeauthoryear{Yu \bgroup \em et al.\egroup }{2012}]{yu2012adaptive}
Jun Yu, Dacheng Tao, and Meng Wang.
\newblock Adaptive hypergraph learning and its application in image classification.
\newblock {\em IEEE Transactions on Image Processing}, 21(7):3262--3272, 2012.

\bibitem[\protect\citeauthoryear{Zhang \bgroup \em et al.\egroup }{2021a}]{zhang2021canonical}
Hengrui Zhang, Qitian Wu, Junchi Yan, David Wipf, and Philip~S Yu.
\newblock From canonical correlation analysis to self-supervised graph neural networks.
\newblock {\em Advances in Neural Information Processing Systems}, 34:76--89, 2021.

\bibitem[\protect\citeauthoryear{Zhang \bgroup \em et al.\egroup }{2021b}]{zhang2021double}
Junwei Zhang, Min Gao, Junliang Yu, Lei Guo, Jundong Li, and Hongzhi Yin.
\newblock Double-scale self-supervised hypergraph learning for group recommendation.
\newblock In {\em Proceedings of the 30th ACM international conference on information \& knowledge management}, pages 2557--2567, 2021.

\bibitem[\protect\citeauthoryear{Zhang \bgroup \em et al.\egroup }{2022}]{zhang2022costa}
Yifei Zhang, Hao Zhu, Zixing Song, Piotr Koniusz, and Irwin King.
\newblock Costa: covariance-preserving feature augmentation for graph contrastive learning.
\newblock In {\em Proceedings of the 28th ACM SIGKDD Conference on Knowledge Discovery and Data Mining}, pages 2524--2534, 2022.

\bibitem[\protect\citeauthoryear{Zhang \bgroup \em et al.\egroup }{2023}]{zhang2023spectral}
Yifei Zhang, Hao Zhu, Zixing Song, Piotr Koniusz, and Irwin King.
\newblock Spectral feature augmentation for graph contrastive learning and beyond.
\newblock In {\em Proceedings of the AAAI Conference on Artificial Intelligence}, volume~37, pages 11289--11297, 2023.

\bibitem[\protect\citeauthoryear{Zhao \bgroup \em et al.\egroup }{2021}]{zhao2021graph}
Han Zhao, Xu~Yang, Zhenru Wang, Erkun Yang, and Cheng Deng.
\newblock Graph debiased contrastive learning with joint representation clustering.
\newblock In {\em IJCAI}, pages 3434--3440, 2021.

\bibitem[\protect\citeauthoryear{Zhou \bgroup \em et al.\egroup }{2006}]{zhou2006learning}
Dengyong Zhou, Jiayuan Huang, and Bernhard Sch{\"o}lkopf.
\newblock Learning with hypergraphs: Clustering, classification, and embedding.
\newblock {\em Advances in neural information processing systems}, 19, 2006.

\bibitem[\protect\citeauthoryear{Zhu \bgroup \em et al.\egroup }{2020}]{zhu2020deep}
Yanqiao Zhu, Yichen Xu, Feng Yu, Qiang Liu, Shu Wu, and Liang Wang.
\newblock Deep graph contrastive representation learning.
\newblock {\em arXiv preprint arXiv:2006.04131}, 2020.

\bibitem[\protect\citeauthoryear{Zhu \bgroup \em et al.\egroup }{2021}]{zhu2021graph}
Yanqiao Zhu, Yichen Xu, Feng Yu, Qiang Liu, Shu Wu, and Liang Wang.
\newblock Graph contrastive learning with adaptive augmentation.
\newblock In {\em Proceedings of the Web Conference 2021}, pages 2069--2080, 2021.

\end{thebibliography}

\end{document}